\documentclass[11pt]{article}
\usepackage{amssymb}

\usepackage{acl}

\usepackage{times}
\usepackage{latexsym}
\usepackage[T1]{fontenc}
\usepackage[utf8]{inputenc}
\usepackage{microtype}
\usepackage{inconsolata}
\usepackage{booktabs}
\usepackage{amsmath}
\usepackage{multirow}
\usepackage{graphicx}
\usepackage{float}

\title{Token Merging for Multilingual Speech Recognition: \\
       A Systematic Study Across Model Scale and Fine-Tuning}
\author{Dylan Luke Holyoak  
\\Geffen Academy at UCLA\\\texttt{dholyoa40@geffenacademy.ucla.edu}}

\begin{document}
\maketitle

\begin{abstract}

Leading multilingual speech recognition models such as Whisper transcribe diverse, low-resource languages without language-specific training but are computationally expensive to deploy. Token merging mitigates this inefficiency by dynamically combining redundant features, shortening the sequence length during inference without requiring retraining the language model. In this paper, we systematically evaluate token merging on the Whisper model family across sixteen diverse languages and three different model sizes. We also test how token merging interacts with fine-tuning (DoRA) on low-resource languages. Our findings show that merging tokens increases computational efficiency with almost no loss in transcription accuracy across most low-resource languages and model sizes, and it works even after the model has been fine-tuned. Our results demonstrate that token merging is a highly practical method for making multilingual speech recognition faster and cheaper to deploy. Code and frozen evaluation results are available at \url{https://github.com/chironk/token-merging-multilingual-asr}.

\end{abstract}
 
\section{Introduction}
Speech models have traditionally been trained on large amounts of
labeled training data, which is only available for a small number of languages. More recently, multilingual automatic speech recognition (ASR) systems have shown that a single model can be trained to transcribe speech in many languages, rather than a separate model per language. Such systems are particularly useful for low-resource languages, where the amount of audio and transcripts available is inadequate to support a per-language model. Recent multilingual systems include weakly supervised models such as MMS \citep{pratap2024scaling} and Whisper \citep{radford2023robust}, and
self-supervised models such as XLS-R \citep{conneau2021xlsr}. These
models demonstrate that multilingual training can be achieved by learning  powerful representations from speech audio alone within a single network. These multilingual models show competitive transcription performance for roughly one hundred to even over a thousand languages.

Among these multilingual ASR models, Whisper \citep{radford2023robust} is the most widely used open-source model, taking speech audio as input and producing transcriptions across roughly one hundred languages without fine-tuning for each language. Although Whisper achieves near-human performance in speech recognition, its audio encoder introduces a significant computational bottleneck. The model is explicitly designed to process fixed-length, 30-second audio inputs, which are encoded into 1,500 tokens and passed through transformer layers. In other words, whether processing a 1-second voice command or a full 30-second monologue, the audio encoder evaluates the 30-second padded sequence. This fixed cost dominates inference for short utterances, creating a major barrier to cost-effective deployment at scale. This inefficiency is especially pronounced in multilingual settings, where average utterance durations can vary significantly across languages.

Token merging mitigates this inefficiency by adaptively combining encoder tokens that share highly similar representations. This operation reduces the sequence length within the model while preserving critical information. Because no retraining is required, the compressed model still transcribes every language it was originally trained on. The technique of token merging originated in computer vision, where soft merging of similar tokens in vision transformers was introduced by ToMe \citep{bolya2023token}, alongside related approaches that prune tokens with learned gating networks \citep{rao2021dynamicvit}, reorganize tokens by attention score \citep{liang2022evit}, halt tokens adaptively per layer \citep{yin2022avit}, or vary depth and width per input \citep{meng2022adavit}.

By contrast, improving the efficiency of speech recognition systems via token merging remains relatively underexplored. Current literature is primarily limited to adjacent token merging for monolingual English RNN-T speech recognition models \citep{li2023atome} and analysis of adjacent-token redundancy in speech language models \citep{xiang2026affinity}. To our knowledge, the technique has not been systematically evaluated on a multilingual ASR model, and key questions remain unanswered. Does it generalize across encoder scales spanning roughly an order of magnitude in parameter count? Does it generalize across typologically diverse languages, including the tonal and low-resource cases for which capacity matters most? And does it remain free after adapter-based fine-tuning, when the decoder has been adapted but the encoder has not?

This paper provides such an evaluation. We apply adjacent token merging, because speech audio inputs differ from images in that the encoder sequence has a temporal order that the decoder cross-attends to monotonically. Accordingly, merging must be restricted to adjacent positions to preserve this alignment. The token merging module is based on non-overlapping pairing on similarity, and is placed intra-block between the attention residual and the feed-forward network. We examine the token merging method using Whisper variants with different model scales, and evaluate on sixteen languages, spanning tonal and non-tonal
phonology, eleven language (sub)families, and four benchmark resource tiers \citep{conneau2023fleurs}.

\paragraph{Contributions.}
The present paper makes three contributions:

\begin{enumerate}
\item \textbf{Cross-lingual evaluation.} We evaluate on sixteen
languages covering tonal and non-tonal phonology, eleven language (sub)families, and low-through high-resource tiers, demonstrating
that the technique generalizes substantially beyond the
English-centric setting of prior work.

\item \textbf{Cross-scale evaluation.} We show that canonical adjacent
token merging causes negligible word-error-rate (WER) drift at three
Whisper encoder scales spanning approximately an
order of magnitude in parameter count, indicating that the technique
is not specific to one model size.

\item \textbf{Composition with fine-tuning.} We show that the technique
composes with DoRA \citep{liu2024dora} decoder fine-tuning on all six
low-resource target languages, and generalizes to ten held-out
languages.
\end{enumerate}

\section{Related Work}

\paragraph{Token reduction in transformers.}
The quadratic complexity of self-attention \citep{vaswani2017attention} with respect to sequence length has motivated extensive research into compressing transformer sequences. Early methods, such as DynamicViT \citep{rao2021dynamicvit}, pruned tokens via learned gating networks. Alternatively, \citet{bolya2023token} demonstrated that bipartite soft \emph{merging}---combining representations rather than discarding them---accelerates vision transformers with negligible accuracy loss and without the need for retraining. In the speech domain, A-ToMe \citep{li2023atome} adapted this idea to an English RNN-T recognizer using a cascade of evenly spaced merge layers to show that token merging also works for ASR models. We adopt their merging schedule and key-based similarity signal, but expand the method to the multilingual encoder-decoder Whisper architecture and explore its composition with parameter-efficient fine-tuning. Furthermore, concurrent research on English LLMs \citep{xiang2026affinity} identifies significant adjacent-token redundancy in deep layers, independently corroborating our structural premise.

\paragraph{Efficient speech recognition.}
Several methods improve speech recognition efficiency by altering the model architecture or weights. For instance, Distil-Whisper \citep{gandhi2023distilWhisper} distills the base model into a 5.8$\times$ faster student. LiteASR \citep{kamahori2025liteasr} employs low-rank factorization to halve the size of Whisper-large-v3 while surpassing Whisper-medium's accuracy. BaldWhisper \citep{sy2025baldWhisper} combines embedding compression with layer merging for deployment. However, these approaches all produce new, retrained models, introducing significant overhead when a single model must serve many languages. Other strategies, such as self-supervised pretraining (e.g., wav2vec 2.0 \citep{baevski2020wav2vec}, XLS-R \citep{conneau2021xlsr}) or utilizing parameter-efficient architectures like the Conformer \citep{gulati2020conformer}, offer alternative methods to lighter encoders but still demand full retraining or per-language fine-tuning. In contrast, token merging fundamentally differs by reducing the input sequence length rather than the parameter count. It operates entirely at inference time on existing Whisper models, requiring zero retraining and no per-language calibration. A complementary family of methods instead reduces encoder cost from the input side, by cropping the input before it reaches the encoder. WhisperX \citep{bain23_interspeech} pre-segments long-form audio with a voice-activity-detection front-end and merges the resulting speech regions back into 30-second chunks aligned on low-activity boundaries, reporting a twelve-fold transcription speedup via batched inference. These input-side savings target the padding cost of Whisper's fixed 30-second window, whereas token merging targets the redundancy that remains inside whatever the front-end forwards.

\paragraph{Multilingual ASR.}
Modern multilingual ASR systems, such as MMS \citep{pratap2024scaling} and XLS-R \citep{conneau2021xlsr}, have scaled speech recognition to hundreds of languages; however, they often rely on self-supervised pretraining and require per-language fine-tuning to transcribe effectively. In contrast, Whisper \citep{radford2023robust} provides highly capable zero-shot transcription without language-specific adaptation. We select Whisper as our primary testbed because of its robust zero-shot capabilities and the public availability of its models across multiple encoder scales, an essential requirement for our cross-scale analysis. To rigorously assess cross-lingual performance, we utilize the FLEURS dataset \citep{conneau2023fleurs}, a standardized benchmark spanning 102 languages. Furthermore, while recent analyses have investigated Whisper's decoder behavior at a fine sub-token granularity across varying resource tiers \citep{liang2025beyondwer}, our work complements this research by targeting the computational efficiency of its encoder.

\paragraph{Parameter-efficient fine-tuning.}
Parameter-efficient fine-tuning mitigates the computational burden of full model retraining by updating only a minimal subset of parameters while the majority remain frozen. Within this paradigm, LoRA \citep{hu2022lora} introduces trainable low-rank update matrices alongside the original weights, whereas DoRA \citep{liu2024dora} builds upon this by decomposing the updates into distinct magnitude and directional components to enhance learning stability. Other specialized designs, such as depth-aware adaptation \citep{xiao2026depthaware}, selectively distribute adapters across encoder layers to optimize low-resource speech recognition. For our study, we apply DoRA exclusively to the Whisper decoder (leaving the encoder completely frozen) to cleanly isolate and investigate the interaction between adapter-based decoder fine-tuning and inference-time encoder token merging.

\section{Method}

\subsection{Whisper Encoder Architecture}
Whisper processes a 30-second log-Mel spectrogram through a two-layer
convolutional stem (stride 1 then stride 2) that projects to a
fixed-length sequence of $T_0 {=} 1500$ tokens. The sequence then
passes through $L$ transformer encoder layers, each applying multi-head
self-attention followed by a position-wise feed-forward network. The
relevant architectural constants are $(L, d, d_{ff}, H, n_{\text{mels}})
{=} (12, 768, 3072, 12, 80)$ for Whisper-small, $(24, 1024, 4096, 16,
80)$ for Whisper-medium, and $(32, 1280, 5120, 20, 128)$ for
Whisper-large-v3, where $L$ is the model layers, $d$ the model dimension, $d_{ff}$ the
feed-forward dimension, $H$ the number of attention heads, and
$n_{\text{mels}}$ the number of mel bins. The encoder output is
consumed by an autoregressive decoder through cross-attention.

\subsection{Token Merging Procedure}
We apply the adjacent-pair token merging procedure of
\citet{bolya2023token} as adapted to speech recognition by
\citet{li2023atome}, with four details specialized to the
encoder-decoder Whisper setting.

\paragraph{Similarity signal.}
For each candidate adjacent pair $(t, t{+}1)$ at a merge layer, we
compute cosine similarity on per-head-averaged key vectors
\begin{equation}
\bar{k}_t = \frac{1}{H} \sum_{h=1}^{H} k_t^{(h)},
\quad k_t^{(h)} \in \mathbb{R}^{d/H},
\label{eq:kbar}
\end{equation}
where $k_t^{(h)}$ is the layer's key projection for head $h$. We use
keys rather than hidden states because they are the representation
actually consulted by attention; per-head averaging reduces the per-pair
similarity to a single scalar in $[-1, 1]$ while remaining faithful to
the similarity structure attention sees.

\paragraph{Placement.}
Merging is applied intra-block, between the multi-head attention
residual update and the layer-norm preceding the feed-forward network.
This is the canonical placement of \citet{bolya2023token} and yields
bit-exact equivalence to the unmerged forward when no pairs are merged,
which we verify empirically (see \S\ref{sec:setup}).

\paragraph{Greedy non-overlapping selection.}
Given a target number of merges $m$ at the current layer, we sort
candidate adjacent pairs by descending cosine similarity and greedily
select the top pairs subject to the constraint that no token
participates in more than one pair. The merge operation replaces each
selected pair $(t, t{+}1)$ with the mean of their hidden representations
and removes the second position, shortening the sequence by exactly
$m$.

\paragraph{Adjacency constraint.}
We restrict merging to adjacent token pairs. This is essential in the
speech setting: encoder tokens correspond to fixed time windows in the
input spectrogram, and the decoder cross-attends in approximately
monotonic temporal order. Merging non-adjacent tokens would fuse
acoustically unrelated time windows and disrupt the alignment on which the
decoder relies.

\subsection{Merge-Layer Schedule}
We adopt the cascade-of-evenly-spaced-layers schedule of
\citet{li2023atome}. Concretely, we apply merging at the fixed subset
of encoder layers determined by
\begin{equation}
\mathcal{M}_L = \{2, 5, 8, 11, \ldots\} \cap \{1, 2, \ldots, L{-}1\},
\label{eq:layers}
\end{equation}
which selects every third layer starting from layer 2 (one-indexed) and
excludes the final encoder layer. This yields $|\mathcal{M}_L| = 4, 8,
10$ merge layers for $L = 12, 24, 32$ respectively. Excluding the final
layer preserves a smoothing buffer before the decoder reads the encoder
output. The per-layer reduction rate $r$ is derived from a global
token reduction ratio (TRR) by
\begin{equation}
r = 1 - (1 - \mathrm{TRR})^{1/|\mathcal{M}_L|},
\label{eq:rate}
\end{equation}
so that the multiplicative effect of $|\mathcal{M}_L|$ merge layers
compounds to the desired global sequence reduction. At each merge
layer the number of merges is $m = \lfloor (T_{\text{pre}} - 1) \cdot r
\rfloor$, where $T_{\text{pre}}$ is the sequence length entering the
layer. We do not tune the merge-layer schedule per language or per
model scale.

\section{Experimental Setup}
\label{sec:setup}

\paragraph{Models.}
We evaluate three Whisper variants drawn from the official OpenAI
releases: Whisper-small (244M parameters), Whisper-medium (769M),
and Whisper-large-v3 (1.55B). Our evaluation spans across small to large-scale ASR models. In addition, we fine-tune Whisper-medium using Weight-Decomposed Low-Rank Adaptation (DoRA) \citep{liu2024dora} across six low-resource languages to investigate the potential interplay between token merging and model fine-tuning. All checkpoints
are pinned to a fixed HuggingFace revision for reproducibility.

\paragraph{Fine-tuning.}
We fine-tune the decoder of the Whisper-medium model only (encoder frozen) using DoRA \citep{liu2024dora} on standard
ASR transcription --- the same task as Whisper's pretraining --- using
six mid/low-resource
languages: Vietnamese, Hausa, Lingala, Tamil, Maltese, and
Javanese. This set of languages is selected to be balanced and diverse: three
tonal (Vietnamese, Hausa, Lingala) and three non-tonal (Tamil,
Maltese, Javanese) languages, six distinct language families
(Austroasiatic, Chadic, Bantu, Dravidian, Semitic, Austronesian),
and baseline WER for these languages spanning 14\% to 89\%.
We use the following DoRA hyperparameters: rank 32, $\alpha {=} 64$, target modules
$\{\text{q\_proj}, \text{k\_proj}, \text{v\_proj},
\text{out\_proj}\}$, learning rate $10^{-5}$, 200-step warmup,
effective batch size 32, 2000 steps. Training mixes the six languages
with temperature ($T{=}0.5$) plus a 10\% English
anchor. We use DoRA to fine-tune the model with a training set including the six mid/low-resource languages as listed above. We then test the model using the hold-out test set for these languages. In addition, we evaluate the model on ten additional languages, including four high-resource languages
(English, French, German, Spanish) and six untrained mid/low-resource
languages (Thai, Swahili, Afrikaans, Icelandic, Welsh, Kazakh).

\paragraph{Data and languages.}
We evaluate on FLEURS \citep{conneau2023fleurs} using the Whisper models with different scales.
Table~\ref{tab:langs} lists the sixteen evaluation languages
covering tonal and non-tonal phonology, eleven language
(sub)families, and four FLEURS resource tiers (high/medium/low/very low). The set comprises four high-resource languages and 12 languages of med/low/very low resources. We sample 264 utterances per language from FLEURS to compute base WER for the test set.

\paragraph{Merging conditions.}
We test token reduction ratios TRR $\in \{0, 0.05, 0.10, 0.20, 0.30,
0.40\}$, where TRR $= 0$ is the baseline without token merging. The merge-layer
schedule is fixed across models per Equations~\ref{eq:layers}
and~\ref{eq:rate}.


\paragraph{Evaluation.}
We report word error rate (WER) on Whisper-normalized transcripts (lower-cased, punctuation stripped, with digits
preserved) and report $\Delta$WER in percentage points (pp) relative to the baseline of the same model without token merging on the same samples. We also compare the model efficiency by comparing the empirical measure of computation time before and after token merging.
\begin{table}[t]
\centering\footnotesize
\setlength{\tabcolsep}{4pt}
\caption{Evaluation languages from FLEURS \citep{conneau2023fleurs}, ordered by resource tier
(H\,=\,High, M\,=\,Medium, L\,=\,Low, VL\,=\,Very Low) and tonal
status (T\,=\,tonal, NT\,=\,non-tonal). Base WER (\%) is from the
Whisper-medium model baseline on $n{=}264$ utterances per language.
}
\label{tab:langs}
\begin{tabular}{llccr}
\toprule
Language & Family & Ton. & Res. & Base WER (\%) \\
\midrule
English     & Germanic       & NT & H  & 5.2   \\
French      & Romance        & NT & H  & 8.8   \\
German      & Germanic       & NT & H  & 6.6   \\
Spanish     & Romance        & NT & H  & 3.5   \\
\midrule
Thai        & Tai-Kadai      & T  & M  & 40.6  \\
Vietnamese  & Austroasiatic  & T  & M  & 14.3  \\
Swahili     & Bantu          & NT & M  & 51.7  \\
Tamil       & Dravidian      & NT & M  & 35.7  \\
\midrule
Hausa       & Chadic         & T  & L  & 88.8  \\
Lingala     & Bantu          & T  & L  & 83.2  \\
Afrikaans   & Germanic       & NT & L  & 45.1  \\
Icelandic   & Germanic       & NT & L  & 49.8  \\
Maltese     & Semitic        & NT & L  & 82.2  \\
Welsh       & Celtic         & NT & L  & 36.9  \\
\midrule
Javanese    & Austronesian   & NT & VL & 67.2  \\
Kazakh      & Turkic         & NT & VL & 53.4  \\
\bottomrule
\end{tabular}
\end{table}

\section{Results}
\label{sec:results}

We organize the results around four empirical questions. Does merging
generalize across typologically diverse languages
(\S\ref{sec:res-lingual})?  Does it compose with adapter-based
decoder fine-tuning (\S\ref{sec:res-ft})? Does it generalize across encoder scales
(\S\ref{sec:res-scale})? And what is the realized
efficiency benefit (\S\ref{sec:efficiency})? All $\Delta$WER values are reported in percentage points relative to the unmerged baseline of the same model on the same samples. Since WER is an error metric, lower error indicates better performance of the ASR model. Hence, negative $\Delta$WER indicates less error (better performance) of the model with token merging than the baseline model. Positive $\Delta$WER suggests that token merging degrades performance (more errors). 

\subsection{Cross-Lingual Robustness}
\label{sec:res-lingual}

\begin{figure}[b!]
\centering
\includegraphics[width=\columnwidth]{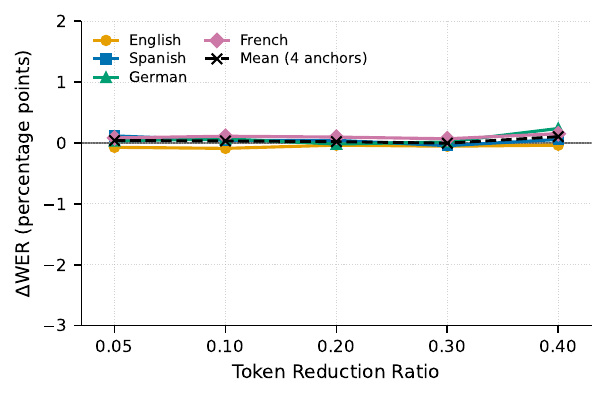}
\vspace{1em}
\includegraphics[width=\columnwidth]{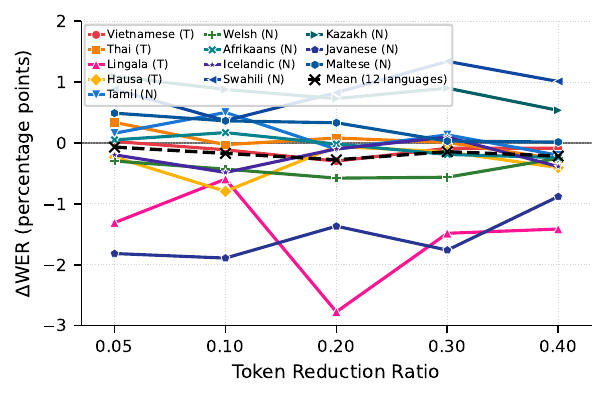}
\caption{$\Delta$WER (percentage points) as a function of token
reduction ratio with the Whisper-medium model for the four
high-resource languages (top) and the 12 main-sweep languages
(bottom).}
\label{fig:highlowres}
\end{figure}

We evaluate adjacent token merging on the Whisper-medium model across sixteen FLEURS languages. Figure~\ref{fig:highlowres}
reports $\Delta$WER as a function of token reduction ratios TRR; per-language values are
reported in Table~\ref{tab:main12} of Appendix~\ref{sec:app-perlang}.

We first examine the twelve med-/low-resource languages that span
tonal (Vietnamese, Thai, Hausa, Lingala) and non-tonal phonology,
Latin-script (Afrikaans, Welsh, Maltese, Icelandic) and
non-Latin-script (Tamil, Thai, Kazakh) writing systems. At $\mathrm{TRR}=0.40$, the cohort-mean $\Delta$WER is $-0.21$ pp, with the largest single-language improvement reaching $-1.41$ pp (Lingala). The aggregate is essentially flat across all five TRRs, varying only between $-0.07$ and $-0.28$ pp. Notably, nine of the twelve languages exhibit a negative $\Delta$WER under token merging, i.e.,\ slightly lower error than the baseline model without token merging. Hence, for mid-/low-resource languages, token merging not only improves computational efficiency but also leaves accuracy intact and in most cases marginally improves it.

To verify that the result is not an artifact of med-/low-resource languages where the baseline without token merging often performs poorly, we additionally evaluate four high-resource anchor languages with \texttt{Whisper-medium} under the same protocol. As shown in the top panel of Figure~\ref{fig:highlowres}, the mean $\Delta$WER at $\mathrm{TRR}=0.40$ is only $+0.10$ pp (English $-0.04$, Spanish $+0.06$, German $+0.24$, French $+0.15$). Despite baseline WERs for these high-resource languages already near the floor (3.5--8.8\%), token merging remains essentially cost-free in accuracy on these languages.

\subsection{Composition with Fine-Tuning}
\label{sec:res-ft}

Table~\ref{tab:ft} compares the Whisper-medium against the DoRA fine-tuned model trained using the six languages, with and without merging at $\mathrm{TRR}=0.40$. Fine-tuning yields substantial baseline gains on four of the six languages, reducing WER for fine-tuned models for Lingala ($83.2 \to 52.2$\%), Javanese ($67.2 \to 51.2$\%), Hausa ($88.8 \to 75.3$\%), and Maltese ($82.2 \to 75.9$\%). Vietnamese ($14.3 \to 14.1$\%) and Tamil ($35.7 \to 36.0$\%) show essentially no change.
\begin{table*}[!ht]
\centering\small
\caption{Whisper-medium vs the DoRA-fine-tuned variant at
TRR\,=\,0.40 on the six trained target languages. For each model we
report the unmerged baseline WER (\%) and the merging effect
$\Delta$@0.40 (pp, vs that model's own baseline). Mean row
(\textbf{bold}) averages all six languages.}
\label{tab:ft}
\begin{tabular}{lrrrr}
\toprule
\multirow{2}{*}{Language}
  & \multicolumn{2}{c}{Whisper-medium}
  & \multicolumn{2}{c}{Whisper-medium DoRA} \\
\cmidrule(lr){2-3}\cmidrule(lr){4-5}
  & Base WER (\%) & $\Delta@0.40$ & Base WER (\%) & $\Delta@0.40$ \\
\midrule
Vietnamese  & 14.3  & $-0.09$ & 14.1  & $+0.17$ \\
Tamil       & 35.7  & $-0.20$ & 36.0  & $-1.41$ \\
Javanese    & 67.2  & $-0.88$ & 51.2  & $-0.41$ \\
Maltese     & 82.2  & $+0.02$ & 75.9  & $+0.13$ \\
Lingala     & 83.2  & $-1.41$ & 52.2  & $+0.59$ \\
Hausa       & 88.8  & $-0.40$ & 75.3  & $-0.84$ \\
\midrule
\textbf{Mean (6)} & \textbf{61.9} & $\mathbf{-0.50}$ & \textbf{50.8} & $\mathbf{-0.29}$ \\
\bottomrule
\end{tabular}
\end{table*}

Applying token merging on top of fine-tuning costs at most $+0.59$ pp on any of the six languages at $\mathrm{TRR}=0.40$ (Lingala), and improves WER on three of these languages. Fine-tuning the multilingual ASR models and token merging are therefore additive in practice: fine-tuning unlocks double-digit WER reductions on the weakest languages (Lingala $-31$, Javanese $-16$, Hausa $-13$ pp), while token merging remains approximately free on the resulting model. The mean $\Delta$WER at $\mathrm{TRR}=0.40$ is $-0.29$ pp on the fine-tuned model versus $-0.50$ pp on the baseline model over the same six languages—i.e.\ merging still reduces WER, just by $0.21$ pp less. 

Beyond the trained six languages used in the fine-tuning, we evaluate the fine-tuned model on ten held-out languages: four high-resource anchor languages (English, French, German, Spanish) and six untrained mid-/low-resource languages (Thai, Swahili, Afrikaans, Icelandic, Welsh, Kazakh). Fine-tuning preserves accuracy on the high-resource languages (mean drift from baseline $+0.02$ pp; max $+0.29$ pp on German) and induces only modest drift on the six untrained languages (mean $+1.77$ pp; max $+3.78$ pp on Kazakh). Token merging at $\mathrm{TRR}=0.40$ remains approximately free across the untrained languages: mean $\Delta$WER $=+0.23$ pp, with the largest single-language excursion $+0.97$ pp (Icelandic). Per-language detail is given in Table~\ref{tab:holdout} of Appendix~\ref{sec:app-perlang}.

\subsection{Cross-Scale Robustness}
\label{sec:res-scale}

\begin{figure*}[t]
\centering
\includegraphics[width=\linewidth]{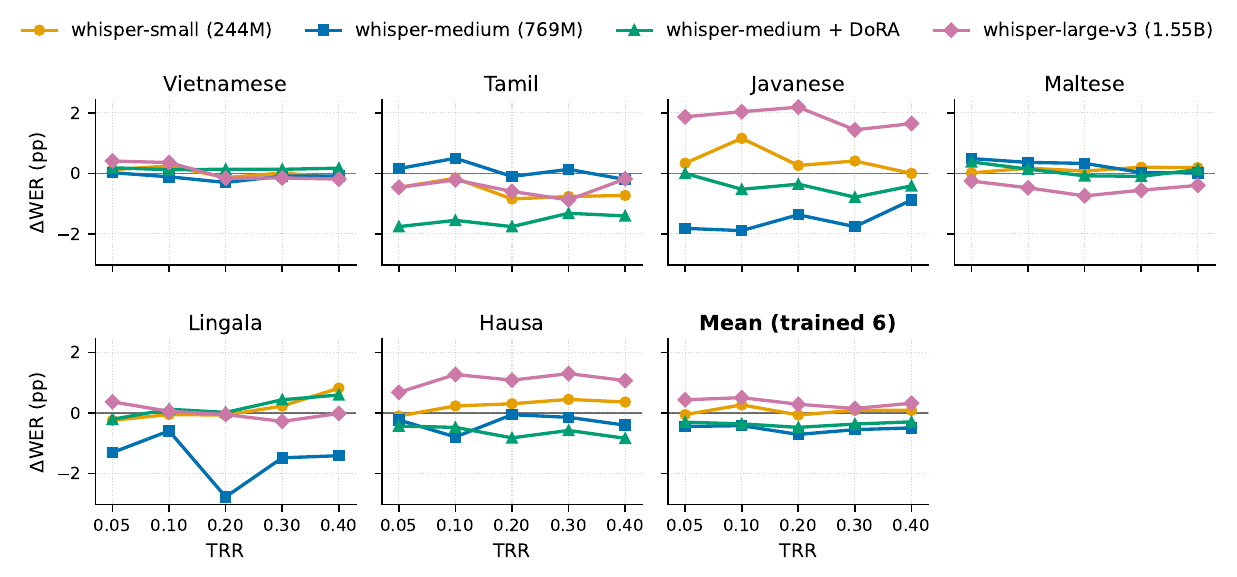}
\caption{Per-language $\Delta$WER (pp) versus token reduction ratios (TRR) on the six languages, relative to each model's baseline without token merging; the
rightmost panel in the bottom row is the mean across the six languages. Four lines per panel:
Whisper-small, Whisper-medium, Whisper-medium with DoRA fine-tuning,
and Whisper-large-v3. The mean stays within $\pm 0.75$ pp at every
TRR; the largest single-language excursion is Javanese on
Whisper-large-v3 ($\leq +2.2$ pp; $+1.65$ pp at TRR\,=\,0.40).}
\label{fig:cross-scale-ft}
\end{figure*}

We next test whether the cross-lingual robustness of token merging reported above persists across model scales. Figure~\ref{fig:cross-scale-ft} plots $\Delta$WER on the six languages at every TRR for three model variants, including Whisper-small (244M), Whisper-medium (769M), and Whisper-large-v3 (1.55B). The six languages, Vietnamese, Tamil, Javanese, Maltese, Lingala, and Hausa, were chosen as representative mid-/low-resource languages spanning a wide range of baseline WER (14\% to 89\% on Whisper-medium). At $\mathrm{TRR}=0.40$, the mean $\Delta$WER across the six languages is $+0.08$ pp for Whisper-small, $-0.50$ pp for Whisper-medium, and $+0.32$ pp for Whisper-large-v3. The magnitude of the mean change stays below half a percentage point at every scale. Table~\ref{tab:ft} reports the results at $\mathrm{TRR}=0.40$ on the six languages; per-language and per-TRR detail behind the figure is given in Table~\ref{tab:detail} of Appendix~\ref{sec:app-perlang}.

Several structural patterns emerge. Vietnamese, which has the lowest baseline WER of the six languages, changes little at any model scale ($|\Delta\mathrm{WER}| < 0.5$ pp at $\mathrm{TRR}=0.40$). Whisper-large-v3 carries the only consistent per-language costs above one pp at $\mathrm{TRR}=0.40$—Javanese ($+1.65$ pp) and Hausa ($+1.07$ pp), neither of which appears at the smaller scales.

\begin{figure*}[t!]
\centering
\includegraphics[width=\linewidth]{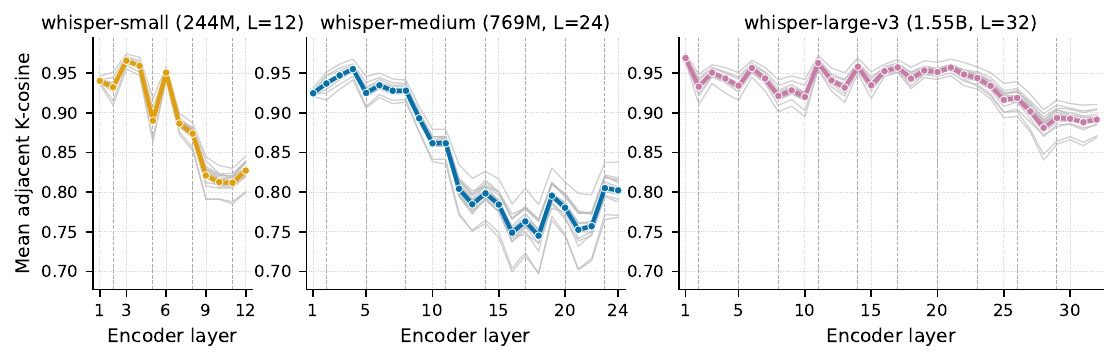}
\caption{Per-layer adjacent K-cosine similarity at the three encoder
scales, with no merging applied ($n{=}264$/lang, 16 languages).
Light-grey lines are per-language profiles, the bold coloured line is
the mean across languages, and dashed verticals mark the encoder layers at which we merge.}
\label{fig:layer-similarity}
\end{figure*}

Figure~\ref{fig:layer-similarity} reveals the structural basis for why token merging is effective for the multilingual ASR models. The per-layer redundancy profile follows the same characteristic shape in every language we test: high adjacent-token similarity in the early and middle encoder layers, tapering in the deepest layers. The curves cluster tightly across languages (cross-language standard deviation $\leq 0.03$ at every layer). This shared, language-agnostic redundancy structure is precisely what token merging exploits.

\subsection{Computational Efficiency}
\label{sec:efficiency}

We measure the computational efficiency using the inference latency of Whisper-medium under 
token merging relative to a baseline model without token merging,
evaluated at token-reduction ratios of $0.20$, $0.30$, and $0.40$. All measurements use CUDA event timers bracketed by device synchronization
rather than host-side wall-clock calls. Each condition is preceded by
discarded warm-up iterations, and a global warm-up drives the GPU to its
sustained clock before timing begins, so the baseline (always measured
first) is not penalized by clock ramp-up. Baseline and token-merged conditions for a given utterance are measured back-to-back so they share
thermal and memory state, and speedups are computed as per-utterance ratios
that are then averaged across utterances, which cancels inter-sample thermal
drift. Results are over
six languages (Vietnamese, Tamil, Javanese, Maltese, Lingala, Hausa), eight
utterances each, on an NVIDIA RTX 3080~Ti (PyTorch~2.6.0,
Transformers~5.5.3, float32, SDPA attention).

We examine encoder speedups relative to the unmerged
baseline, which is a stable $\approx\!94$\,ms across all six languages. Token merging yields a monotonic encoder speedup that grows with the
reduction ratio, reaching $\mathbf{1.27\times}$ at $\mathrm{TRR}=0.40$
(encoder time $94 \rightarrow 74$\,ms). This result is highly consistent
across languages (per-language range $1.26$--$1.27\times$), reflecting that
encoder cost is governed by the merge schedule rather than by linguistic
content. Hence, token merging delivers a reliable $1.27\times$ encoder speedup at
$\mathrm{TRR}=0.40$ with negligible cross-language variance. The encoder speedup with token merging reaches $\mathbf{1.18\times}$ at $\mathrm{TRR}=0.30$ and $\mathbf{1.10\times}$ at $\mathrm{TRR}=0.20$. Because the savings are encoder-localized, the largest end-to-end gains are expected in deployment settings dominated by encoder cost (e.g., large batch sizes or short transcripts).



\section{Discussion}

\paragraph{Fine-tuning unlocks accuracy and composes with merging.}
DoRA fine-tuning on our six target languages yields substantial baseline improvements for four of them: Lingala (-31 pp), Javanese (-16 pp), Hausa (-13 pp), and Maltese (-6 pp), while Vietnamese and Tamil remain essentially unchanged. Crucially, applying token merging on top of this fine-tuning induces a maximum WER degradation of just +0.59 pp across the trained languages at a token reduction ratio (TRR) of 0.40, and at most +0.97 pp across the ten held-out languages. These two efficiency techniques are therefore highly complementary: the DoRA adapter unlocks massive accuracy gains for languages that traditionally perform poorly in multilingual ASR models, while token merging reduces inference costs almost for free on the resulting model—generalizing seamlessly across both seen and unseen languages. 

\paragraph{When to apply token merging.}
Our findings indicate that deployment recommendations are safely decoupled from the target language. The same fixed merging schedule and TRR setting hold across all sixteen evaluated languages, as well as across both the stock and fine-tuned checkpoints, eliminating the need for language-specific or cohort-specific recalibration. Setting TRR = 0.40 maintains a strict 1.5 pp WER budget across every language in our main simulation and all high-resource anchor languages, delivering a reliable 1.27x encoder speedup. For tighter accuracy constraints, lower TRR values trade compute savings for an additional safety margin. Consequently, practitioners can establish the compute-accuracy trade-off a single time at deployment based entirely on their global WER budget, rather than relying on granular, language-by-language fine-tuning. 


\section{Conclusion}

In this work, we presented a systematic evaluation of adjacent token merging for multilingual Whisper models. Our study encompassed three encoder scales (spanning an order of magnitude in parameter count), sixteen typologically diverse languages, and decoder-side DoRA fine-tuning. We demonstrated that adjacent token merging induces negligible word error rate (WER) degradation across all tested model sizes and, in low-resource language scenarios, even marginally improves accuracy for some languages. Furthermore, the technique composes seamlessly with adapter-based fine-tuning across both target and held-out languages. At the most aggressive reduction setting, merging eliminates roughly a quarter of the theoretical encoder compute, which translates to a reliable empirical speedup. Consequently, token merging serves as a highly practical, training-free, and post-hoc efficiency method for deploying multilingual ASR systems.


\section*{Limitations}
We acknowledge two primary limitations to our study. First, we employ a static merging schedule across all languages and model scales, so a single conservative rate has to cover the hardest clip in the batch even when most clips could tolerate more aggressive merging. A natural next step is to train a small neural network that takes the clip's audio features as input and outputs a reduction rate for each merge layer, so every clip gets its own schedule instead of the shared one used here. This network can be trained by comparing the transcripts Whisper produces under its predicted rates against the transcripts the un-merged Whisper produces, rewarding rates that merge aggressively while leaving the transcript unchanged. Furthermore, our fine-tuning composition experiments rely exclusively on the DoRA approach; it remains to be tested whether this additive performance result extends to other parameter-efficient adaptation methods, such as prefix tuning or alternative adapter architectures.  

Second, while our held-out evaluation spans ten languages across diverse resource tiers, this subset represents only a fraction of Whisper's complete pretraining corpus. Consequently, our findings should be interpreted as strong supporting evidence of broad generalization rather than a conclusive proof. Validating the observed trade-off between accuracy and computation speed on a wider array of typologically distant languages, particularly those severely underrepresented in the original pretraining data, would further support these claims.

\bibliography{custom}

\onecolumn
\appendix

\section{Theoretical FLOP Derivation}
\label{sec:app-flops}

The encoder-FLOP figures reported in \S\ref{sec:efficiency}
(24.3--25.1\% reduction at TRR\,=\,0.40 across model scales) are
computed by simulating the integer per-layer arithmetic ($m =
\lfloor (T_{\text{pre}}{-}1) \cdot r \rfloor$) and summing, at each
encoder layer, attention FLOPs at the pre-merge length and
feed-forward FLOPs at the post-merge length:
\begin{align}
\text{Attn}(T) &= 8 T d^{2} + 4 T^{2} d, \nonumber \\
\text{FFN}(T)  &= 4 T d \cdot d_{ff}.
\label{eq:flops}
\end{align}
The fixed cost of the two-layer convolutional stem is included in
both the merged and baseline totals so that the reported reduction
is honest with respect to the encoder as a whole. Layer-norm
contributions are below 0.1\% of layer FLOPs and are omitted per the
standard convention \citep{bolya2023token}.

The gap between the nominal sequence reduction (40\%) and the
realised FLOP reduction ($\sim 25\%$) at TRR\,=\,0.40 is structural.
Only a subset of layers ($|\mathcal{M}_L|$ out of $L$) perform
merging, layers before the first merge always operate at the full
length $T_0$, and the feed-forward term scales linearly rather than
quadratically in $T$, dominating the per-layer cost at our
$d_{ff}/d$ ratio of 4.

\section{Per-Language Tables}
\label{sec:app-perlang}

\begin{table}[H]
\centering\small
\setlength{\tabcolsep}{4pt}
\caption{12-language main sweep on Whisper-medium with merging
at the schedule of Eq.~\ref{eq:layers}. Columns 0.05--0.40 are
$\Delta$WER (pp) at each token reduction ratio; baseline WER (\%) at
TRR\,=\,0 shown alongside. The mean row (\textbf{bold}) averages all
twelve languages.}
\label{tab:main12}
\begin{tabular}{lrrrrrr}
\toprule
Language & Base WER (\%) & 0.05 & 0.10 & 0.20 & 0.30 & 0.40 \\
\midrule
Vietnamese  & 14.3  & $+0.02$ & $-0.11$ & $-0.30$ & $-0.09$ & $-0.09$ \\
Tamil       & 35.7  & $+0.16$ & $+0.50$ & $-0.10$ & $+0.13$ & $-0.20$ \\
Welsh       & 36.9  & $-0.30$ & $-0.43$ & $-0.58$ & $-0.56$ & $-0.25$ \\
Thai        & 40.6  & $+0.34$ & $-0.03$ & $+0.08$ & $+0.01$ & $-0.27$ \\
Afrikaans   & 45.1  & $+0.05$ & $+0.17$ & $-0.02$ & $-0.19$ & $-0.24$ \\
Icelandic   & 49.8  & $-0.19$ & $-0.48$ & $-0.10$ & $+0.10$ & $-0.39$ \\
Swahili     & 51.7  & $+0.88$ & $+0.37$ & $+0.83$ & $+1.34$ & $+1.01$ \\
Kazakh      & 53.4  & $+1.09$ & $+0.88$ & $+0.73$ & $+0.90$ & $+0.54$ \\
Javanese    & 67.2  & $-1.82$ & $-1.89$ & $-1.37$ & $-1.76$ & $-0.88$ \\
Maltese     & 82.2  & $+0.49$ & $+0.36$ & $+0.33$ & $+0.03$ & $+0.02$ \\
Lingala     & 83.2  & $-1.31$ & $-0.59$ & $-2.78$ & $-1.48$ & $-1.41$ \\
Hausa       & 88.8  & $-0.23$ & $-0.80$ & $-0.06$ & $-0.14$ & $-0.40$ \\
\midrule
\textbf{Mean (12)} & \textbf{54.1} & $\mathbf{-0.07}$ & $\mathbf{-0.17}$ & $\mathbf{-0.28}$ & $\mathbf{-0.14}$ & $\mathbf{-0.21}$ \\
\bottomrule
\end{tabular}
\end{table}

\begin{table}[H]
\centering\small
\caption{Per-language $\Delta$WER (pp) on the six trained languages
at each TRR, for each of the three stock model scales plus the DoRA
fine-tuned medium variant. Baseline (TRR\,=\,0) WER (\%) is shown in
parentheses next to each language. The mean row (\textbf{bold}) averages
over the six trained languages.}
\label{tab:detail}
\begin{tabular}{llrrrrr}
\toprule
Model & Language & 0.05 & 0.10 & 0.20 & 0.30 & 0.40 \\
\midrule
\multirow{7}{*}{Whisper-small}
  & Vietnamese (21.9) & $+0.12$ & $+0.25$ & $-0.14$ & $+0.01$ & $-0.14$ \\
  & Tamil (43.5)      & $-0.46$ & $-0.16$ & $-0.84$ & $-0.76$ & $-0.73$ \\
  & Javanese (82.2)   & $+0.34$ & $+1.16$ & $+0.26$ & $+0.41$ & $\hphantom{+}0.00$ \\
  & Maltese (89.3)    & $+0.02$ & $+0.17$ & $+0.08$ & $+0.21$ & $+0.19$ \\
  & Lingala (90.4)    & $-0.24$ & $-0.05$ & $-0.07$ & $+0.23$ & $+0.82$ \\
  & Hausa (88.4)      & $-0.10$ & $+0.23$ & $+0.30$ & $+0.45$ & $+0.36$ \\
  & \textbf{Mean}     & $\mathbf{-0.06}$ & $\mathbf{+0.27}$ & $\mathbf{-0.07}$ & $\mathbf{+0.09}$ & $\mathbf{+0.08}$ \\
\midrule
\multirow{7}{*}{Whisper-medium}
  & Vietnamese (14.3) & $+0.02$ & $-0.11$ & $-0.30$ & $-0.09$ & $-0.09$ \\
  & Tamil (35.7)      & $+0.16$ & $+0.50$ & $-0.10$ & $+0.13$ & $-0.20$ \\
  & Javanese (67.2)   & $-1.82$ & $-1.89$ & $-1.37$ & $-1.76$ & $-0.88$ \\
  & Maltese (82.2)    & $+0.49$ & $+0.36$ & $+0.33$ & $+0.03$ & $+0.02$ \\
  & Lingala (83.2)    & $-1.31$ & $-0.59$ & $-2.78$ & $-1.48$ & $-1.41$ \\
  & Hausa (88.8)      & $-0.23$ & $-0.80$ & $-0.06$ & $-0.14$ & $-0.40$ \\
  & \textbf{Mean}     & $\mathbf{-0.45}$ & $\mathbf{-0.42}$ & $\mathbf{-0.71}$ & $\mathbf{-0.55}$ & $\mathbf{-0.50}$ \\
\midrule
\multirow{7}{*}{med.\,DoRA}
  & Vietnamese (14.1) & $+0.19$ & $+0.12$ & $+0.14$ & $+0.14$ & $+0.17$ \\
  & Tamil (36.0)      & $-1.76$ & $-1.55$ & $-1.76$ & $-1.32$ & $-1.41$ \\
  & Javanese (51.2)   & $\hphantom{+}0.00$ & $-0.52$ & $-0.36$ & $-0.79$ & $-0.41$ \\
  & Maltese (75.9)    & $+0.38$ & $+0.14$ & $-0.08$ & $-0.10$ & $+0.13$ \\
  & Lingala (52.2)    & $-0.21$ & $+0.12$ & $+0.02$ & $+0.44$ & $+0.59$ \\
  & Hausa (75.3)      & $-0.43$ & $-0.48$ & $-0.82$ & $-0.58$ & $-0.84$ \\
  & \textbf{Mean}     & $\mathbf{-0.31}$ & $\mathbf{-0.36}$ & $\mathbf{-0.48}$ & $\mathbf{-0.37}$ & $\mathbf{-0.29}$ \\
\midrule
\multirow{7}{*}{large-v3}
  & Vietnamese (10.8) & $+0.41$ & $+0.36$ & $-0.16$ & $-0.15$ & $-0.19$ \\
  & Tamil (32.1)      & $-0.46$ & $-0.21$ & $-0.60$ & $-0.88$ & $-0.18$ \\
  & Javanese (60.1)   & $+1.87$ & $+2.04$ & $+2.19$ & $+1.44$ & $+1.65$ \\
  & Maltese (68.1)    & $-0.25$ & $-0.47$ & $-0.74$ & $-0.55$ & $-0.40$ \\
  & Lingala (70.1)    & $+0.37$ & $+0.05$ & $-0.05$ & $-0.28$ & $-0.02$ \\
  & Hausa (83.5)      & $+0.68$ & $+1.27$ & $+1.08$ & $+1.30$ & $+1.07$ \\
  & \textbf{Mean}     & $\mathbf{+0.44}$ & $\mathbf{+0.51}$ & $\mathbf{+0.29}$ & $\mathbf{+0.15}$ & $\mathbf{+0.32}$ \\
\bottomrule
\end{tabular}
\end{table}

\begin{table}[H]
\centering\small
\setlength{\tabcolsep}{4pt}
\caption{Held-out cohort merging sweep on the DoRA-fine-tuned
Whisper-medium model ($n{=}264$/lang). Columns 0.05--0.40 are
$\Delta$WER (pp) at each TRR vs the DoRA-fine-tuned unmerged
baseline; ``Base WER (\%)'' is the DoRA unmerged WER at TRR\,=\,0.
Four high-resource anchors above the rule; six untrained
mid/low-resource languages below. Mean rows (\textbf{bold}) report
each tier separately and the cohort overall.}
\label{tab:holdout}
\begin{tabular}{lrrrrrr}
\toprule
Language & Base WER (\%) & 0.05 & 0.10 & 0.20 & 0.30 & 0.40 \\
\midrule
English   & 4.9  & $+0.05$ & $+0.07$ & $+0.11$ & $+0.12$ & $+0.05$ \\
French    & 9.0  & $+0.11$ & $+0.06$ & $+0.01$ & $+0.01$ & $+0.10$ \\
German    & 6.9  & $+0.15$ & $+0.15$ & $+0.15$ & $+0.10$ & $+0.17$ \\
Spanish   & 3.4  & $+0.10$ & $+0.15$ & $+0.13$ & $+0.15$ & $+0.10$ \\
\midrule
\textbf{Mean (4 anchors)} & \textbf{6.1} & $\mathbf{+0.10}$ & $\mathbf{+0.11}$ & $\mathbf{+0.10}$ & $\mathbf{+0.10}$ & $\mathbf{+0.11}$ \\
\midrule
Thai      & 42.4 & $-0.06$ & $-0.26$ & $+0.16$ & $-0.37$ & $-0.08$ \\
Welsh     & 38.3 & $+0.31$ & $+0.43$ & $+0.33$ & $+0.55$ & $+0.25$ \\
Afrikaans & 44.8 & $+0.39$ & $+0.38$ & $+0.32$ & $+0.12$ & $+0.15$ \\
Icelandic & 51.2 & $+1.06$ & $+1.26$ & $+1.06$ & $+1.84$ & $+0.97$ \\
Swahili   & 54.2 & $+0.57$ & $+0.40$ & $+0.59$ & $+0.72$ & $+0.46$ \\
Kazakh    & 57.2 & $+0.75$ & $+0.41$ & $+0.52$ & $+0.30$ & $+0.13$ \\
\midrule
\textbf{Mean (6 untrained)} & \textbf{48.0} & $\mathbf{+0.50}$ & $\mathbf{+0.44}$ & $\mathbf{+0.50}$ & $\mathbf{+0.53}$ & $\mathbf{+0.31}$ \\
\midrule
\textbf{Mean (all 10)} & \textbf{31.2} & $\mathbf{+0.34}$ & $\mathbf{+0.31}$ & $\mathbf{+0.34}$ & $\mathbf{+0.35}$ & $\mathbf{+0.23}$ \\
\bottomrule
\end{tabular}
\end{table}

\end{document}